\documentclass[letterpaper, 10 pt, conference]{ieeeconf}  

\IEEEoverridecommandlockouts                              

\usepackage{graphicx} 
\usepackage{amsmath} 
\usepackage{amssymb}  
\usepackage{hyperref}
\usepackage{todonotes}
\usepackage{subfigure}
\usepackage{bm}

\title{\LARGE \bf
Quantitative Depth Quality Assessment of RGBD Cameras At Close Range Using 3D Printed Fixtures
}

\author{Michele Pratusevich, Jason Chrisos, and Shreyas Aditya
\thanks{Root AI Inc, 444 Somerville Ave, Somerville, MA 02143, USA.
        {\tt\small \{mprat, jchrisos, sadiyta\}@root-ai.com}}%
}

\begin{document}

\maketitle
\thispagestyle{empty}
\pagestyle{empty}

\begin{abstract}

Mobile robots that manipulate their environments require high-accuracy scene understanding at close range. Typically this understanding is achieved with RGBD cameras, but the evaluation process for selecting an appropriate RGBD camera for the application is minimally quantitative. Limited manufacturer-published metrics do not translate to observed quality in real-world cluttered environments, since quality is application-specific. To bridge the gap, we present a method for quantitatively measuring depth quality using a set of extendable 3D printed fixtures that approximate real-world conditions. By framing depth quality as point cloud density and root mean square error (RMSE) from a known geometry, we present a method that is extendable by other system integrators for custom environments. We show a comparison of 3 cameras and present a case study for camera selection, provide reference meshes and analysis code\footnote{Meshes and code are available at: \url{http://github.com/Root-AI/depth-quality}}, and discuss further extensions.

\end{abstract}

\section{INTRODUCTION}

Fine or delicate robotic manipulation tasks require high quality, high accuracy, and high speed 3D scene understanding at close range. Typically, high-level understanding algorithms are built on top of an RGBD camera sensor stream that gives 3D world positions for every point in the camera view. Camera manufacturers such as Intel\textsuperscript{\textregistered} Realsense\textsuperscript{\texttrademark} \cite{realsense_d400_datasheet}, StereoLabs \cite{StereoLabs:ZedMini}, Orbbec \cite{orbbec_astra}, ASUS \cite{noauthor_xtion} and others publish limited metrics about the density, accuracy, and quality of the depth maps produced by their cameras. Depth accuracy (through root mean square error (RMSE)) is typically measured against flat plane targets, which does not accurately capture the geometry of real-world scenes.

\begin{figure}[tpb]
\centering
\parbox{3.4in}{
\includegraphics[width=0.32\linewidth]{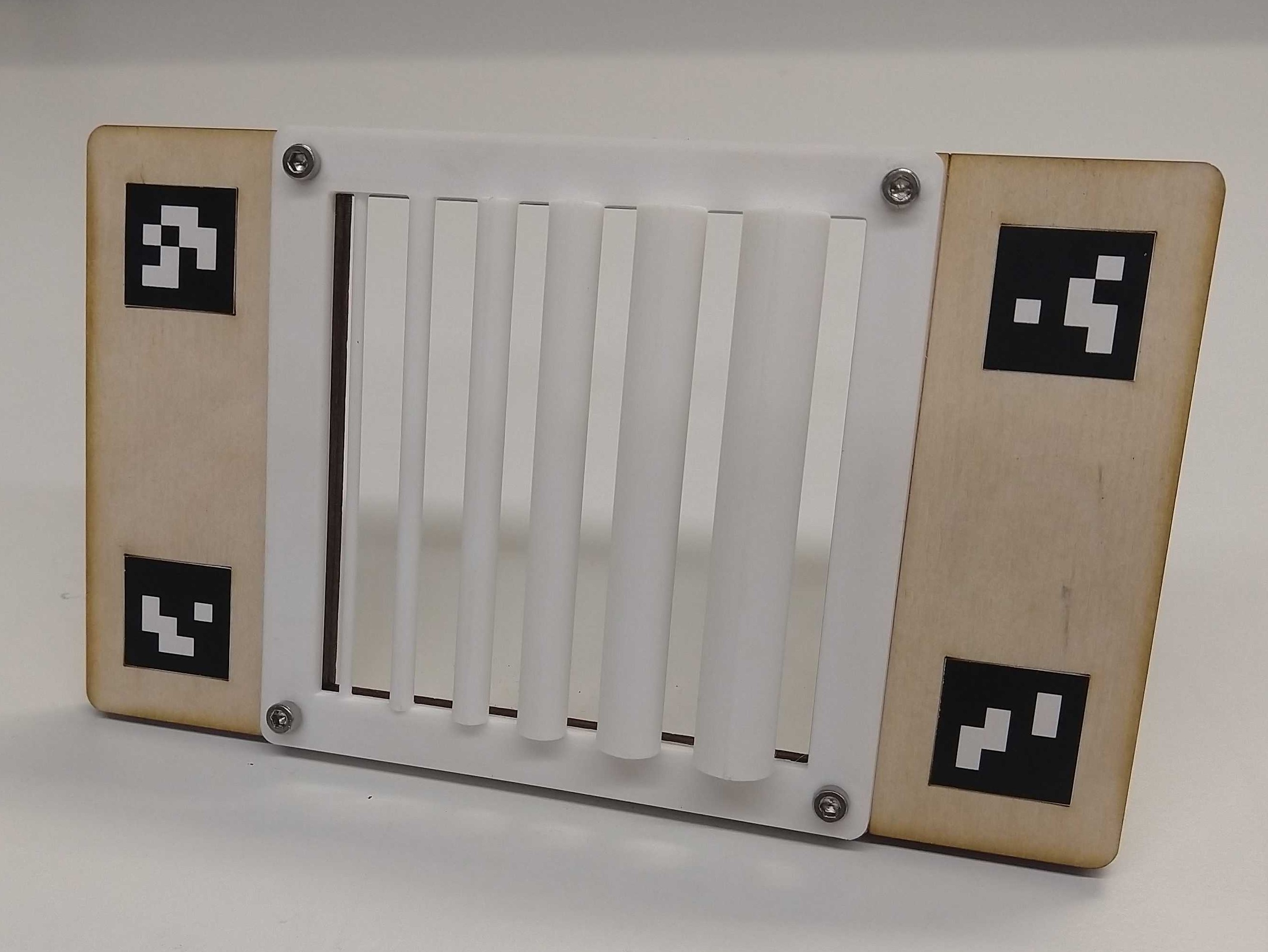}
\includegraphics[width=0.32\linewidth]{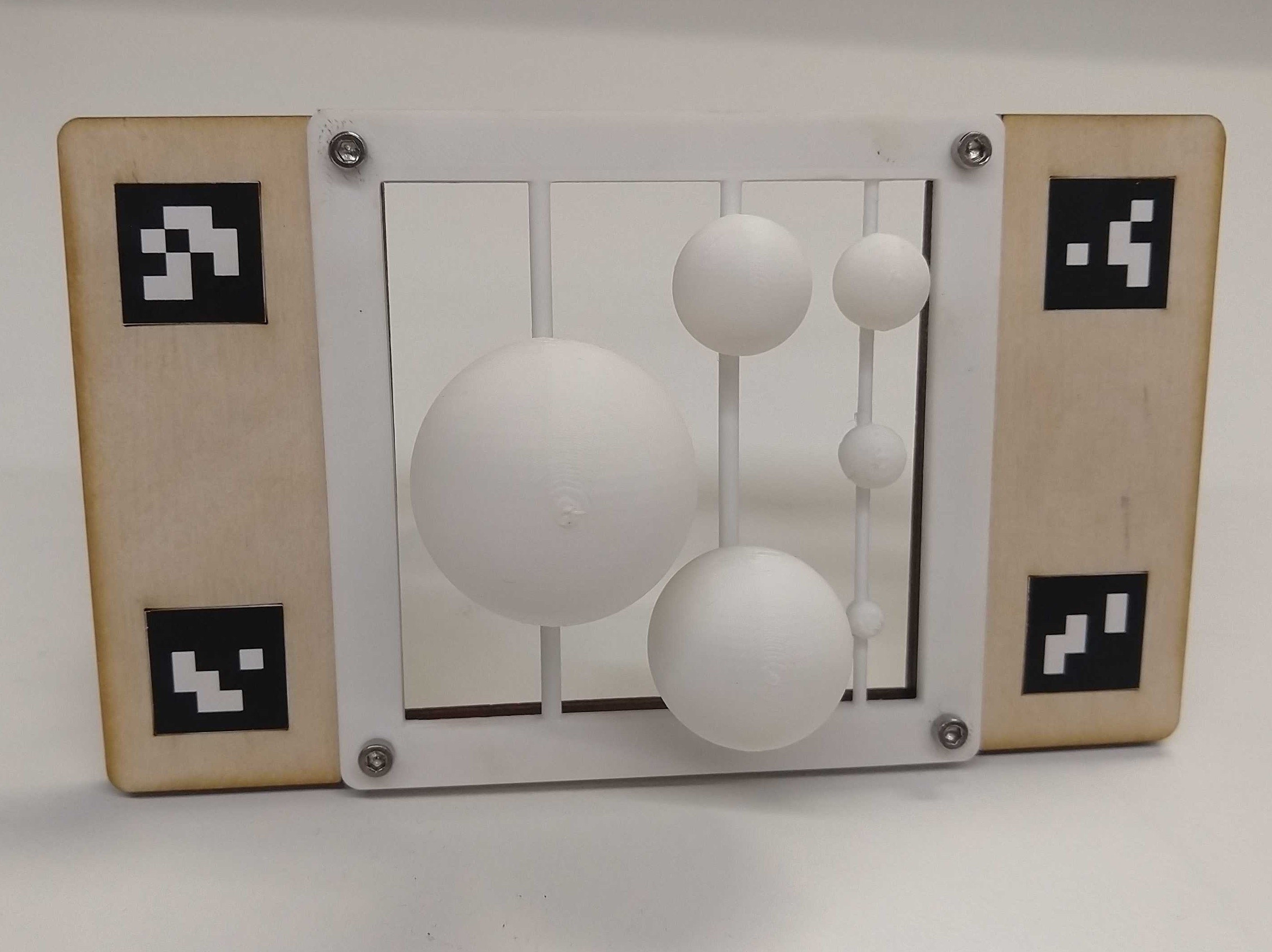}
\includegraphics[width=0.32\linewidth]{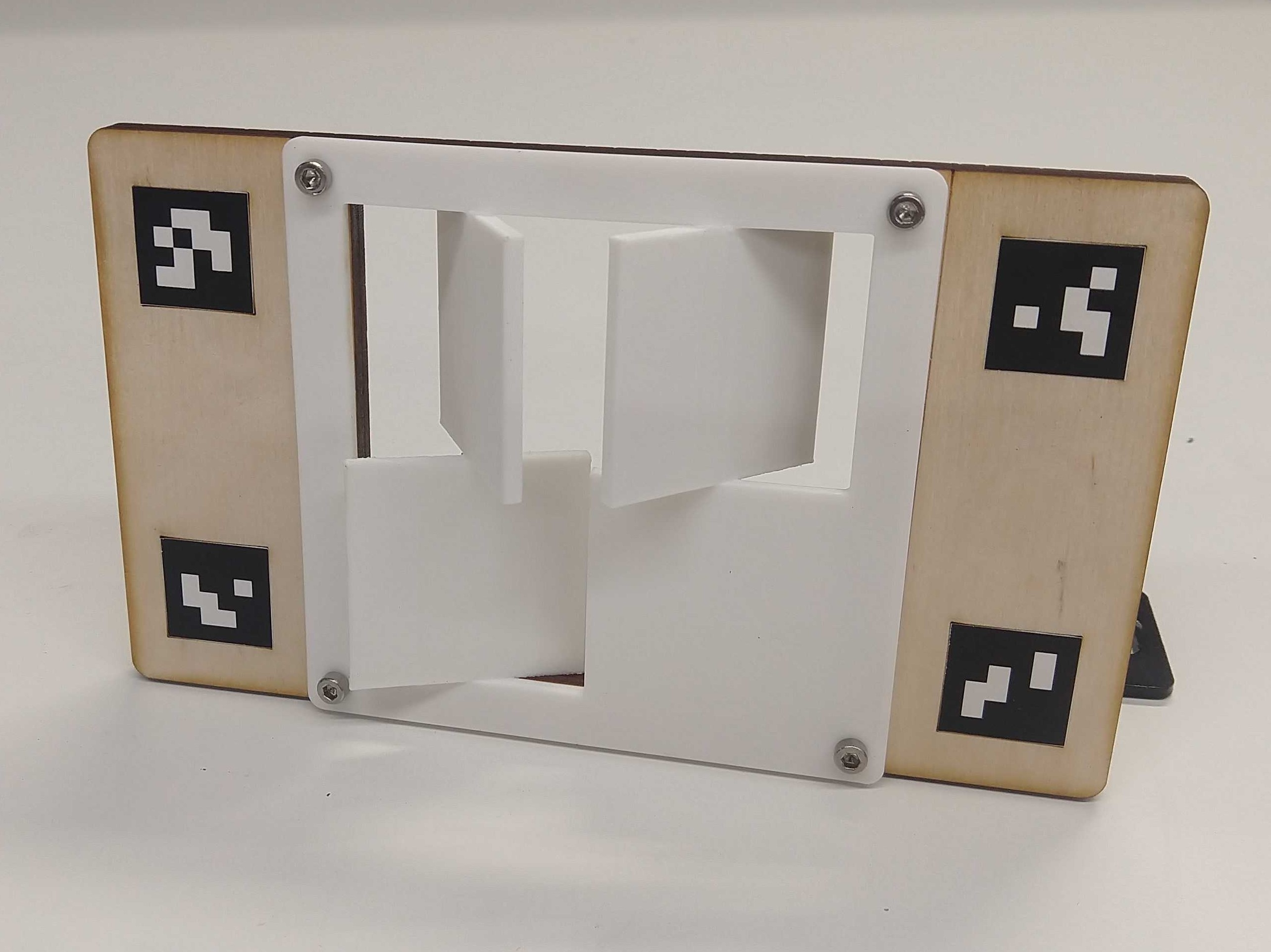}
}
\parbox{3.4in}{
\includegraphics[width=0.32\linewidth]{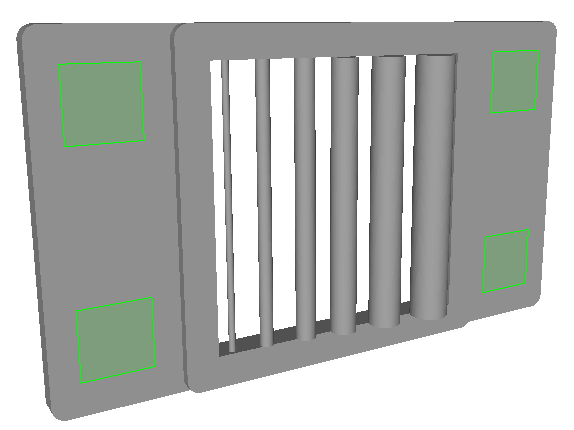}
\includegraphics[width=0.32\linewidth]{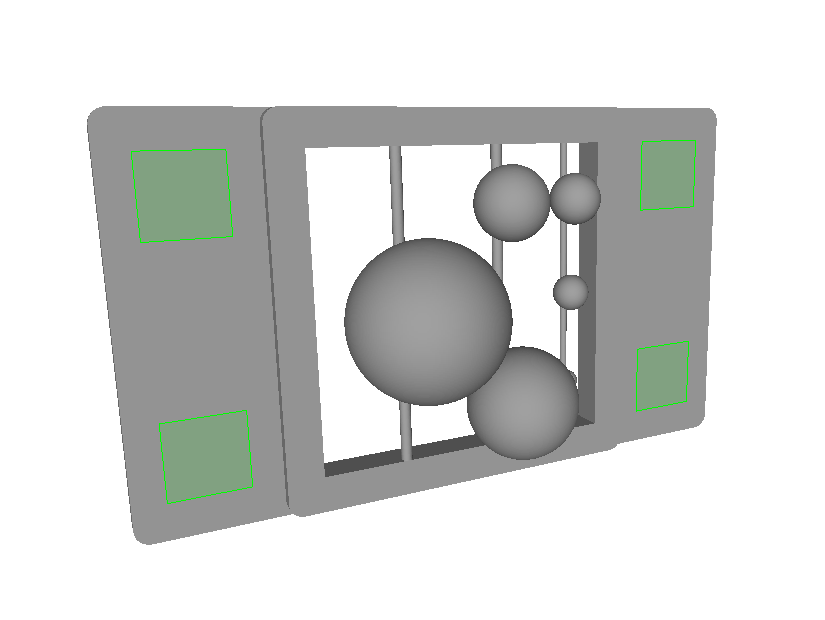}
\includegraphics[width=0.32\linewidth]{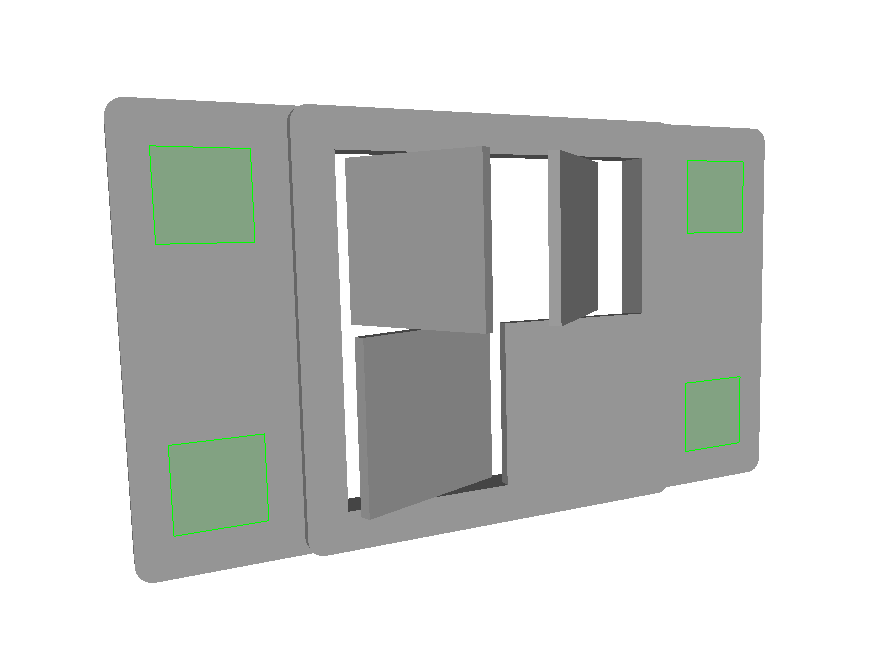}
}
\caption{The 3 test fixtures and their visualized reference meshes.}
\label{fig:fixtures}
\end{figure}

Instead, we measure depth accuracy through RMSE but use targets of known geometry that better approximates typical structures seen in manipulation environments, as shown in Fig. \ref{fig:fixtures}. We discuss the fixture design and manufacturing process, the specific definition we use for defining depth quality (a combination of RMSE and density), and discuss how this procedure was used at Root AI for camera selection.

Robotic manipulation systems are typically designed with a specific environment and target operating distance in mind. The modular fixture design gives integrators the flexibility to design custom test patterns specific to their environments and test distances while still using the same base fixture, alignment and evaluation procedure, code, and metrics. By leveraging 3D printing to model simple environment patterns, pointclouds captured from different cameras, settings, or algorithms can be compared against the same ground truth. This method can be applied to all kinds of RGBD cameras, including structured light, time-of-flight (ToF), passive stereo, etc.

\section{DEPTH QUALITY}

There is no standard definition of depth quality for an end-to-end stereo camera system. Stereo correspondence algorithms are benchmarked against standard datasets like the Middlebury \cite{scharstein_taxonomy_2001} or KITTI \cite{geiger_are_2012} stereo datasets, where the inputs are pairs of stereo images, and the evaluation metrics are RMSE, the number of bad (invalid) pixels, and the prediction error, measured across different classes of pixels (e.g. occluded, textured, etc.). However, these metrics and datasets test algorithms, not the performance of cameras in real-world environments.

Intel\textsuperscript{\textregistered} Realsense\textsuperscript{\texttrademark} \cite{keselman_intelr_2017} do explain their testing methodology for determining depth camera error, testing against flat, white walls and measuring the RMSE of the observed values against the best-fit plane. This technique is simple and straightforward - offices are filled with flat white walls, and best-fit plane algorithms can be checked with straightforward measurements. Haggag et. al \cite{haggag_measuring_2013} and Wasenm\"{u}ller and Stricker \cite{chen_comparison_2017} also use planar targets to measure accuracy using pixel depth error. The simplicity of using a flat wall target allows testing of multiple factors (temperature, target distance, image location), which provides a comprehensive understanding of the RGBD sensor limitations. However, flat white walls rarely appear at close range in real-world environments where robotic manipulation systems operate, especially outdoors.

Lachat et. al. \cite{lachat_assessment_2015} measure accuracy of the Kinect v1 and v2 RGBD sensors using a sandstone balustrate fragment. Accuracy was measured by the density of points in a reconstructed pointcloud and the RMSE of observed values. The ground truth mesh of the fragment was constructed from a sub-millimeter scanner, and multi-image registration and reconstruction was applied to the observed pointclouds. Our method improves on the method used here by producing from scratch objects of known geometries, reducing the need for expensive scanning technologies to generate ground-truth. 

In this work, we define depth quality with two metrics: RMSE and density against known targets that approximate real-world scenes, discussed in Section \ref{sec:metric}. Additionally, our method operates on a single frame or capture of the object, rather than wrapping a 3D reconstruction method (and any associated error from it) into the accuracy estimation. Most real-world robotics manipulation pipelines operate in moving environments, motivating the need to choose an RGBD camera that provides a high-accuracy depth image from a single frame and viewpoint without requiring multi-view 3D reconstruction.

\section{METHODOLOGY}

We define depth accuracy through pixel density and RMSE of a known test fixture. To evaluate this metric, we (1) manufacture a test fixture with known geometry that can be accurately localized in 3D space (Section \ref{sec:fixture}), (2) capture data using an RGBD camera, (3) align the observed data to the reference mesh (Section \ref{sec:alignment}), and (4) compute our accuracy measure (Section \ref{sec:metric}).

\label{sec:fixture}

\begin{figure}[tpb]
\centering
\parbox{3in}{
\includegraphics[width=\linewidth]{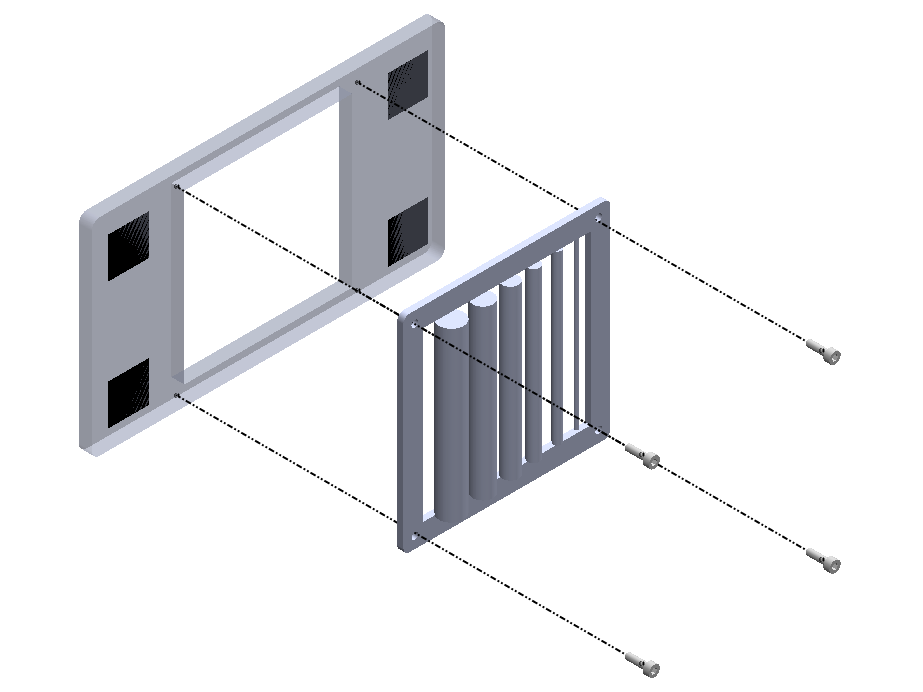}
}
\caption{The test fixture assembly - the backplate for localization, the pattern plate for attaching the test pattern to the backplate, and the test pattern for analysis.}
\label{fig:assembly}
\end{figure}

Manipulation environments at close range are composed of a base set of features that can be reproduced with CAD modeling software and a consumer-grade fused deposition modeling (FDM) 3D printer. For our test fixtures, we selected cylinders, spheres, and planes, of differing sizes and angles, as shown in Fig. \ref{fig:fixtures}. These 3 fixtures are versatile decompositions of objects commonly observed in manipulation environments, and the varying sizes and angles give a more detailed understanding of what scene properties the camera performs worst at. Additionally, the angled plate fixture in particular is designed to evaluate depth quality on textureless targets that are historically challenging for depth cameras \cite{seitz_comparison_2006}.

\begin{figure*}[tpb]
\centering
\parbox{6.8in}{
\subfigure[]{	
\includegraphics[width=0.32\linewidth]{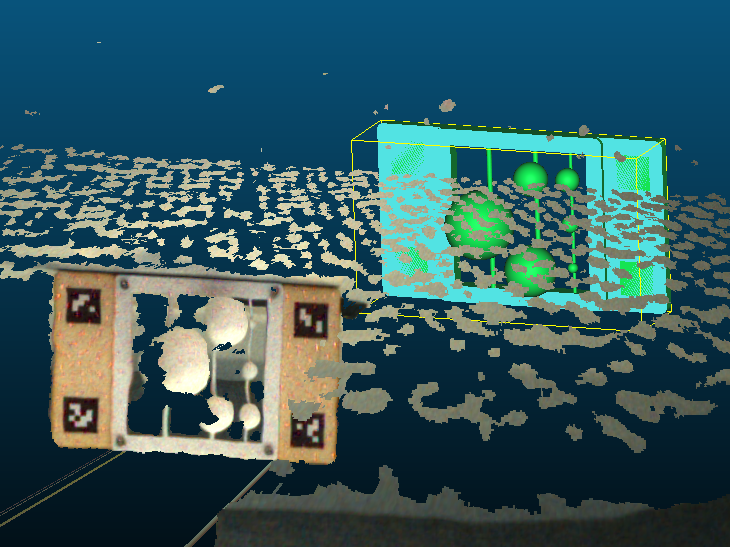}
}
\subfigure[]{	
\includegraphics[width=0.32\linewidth]{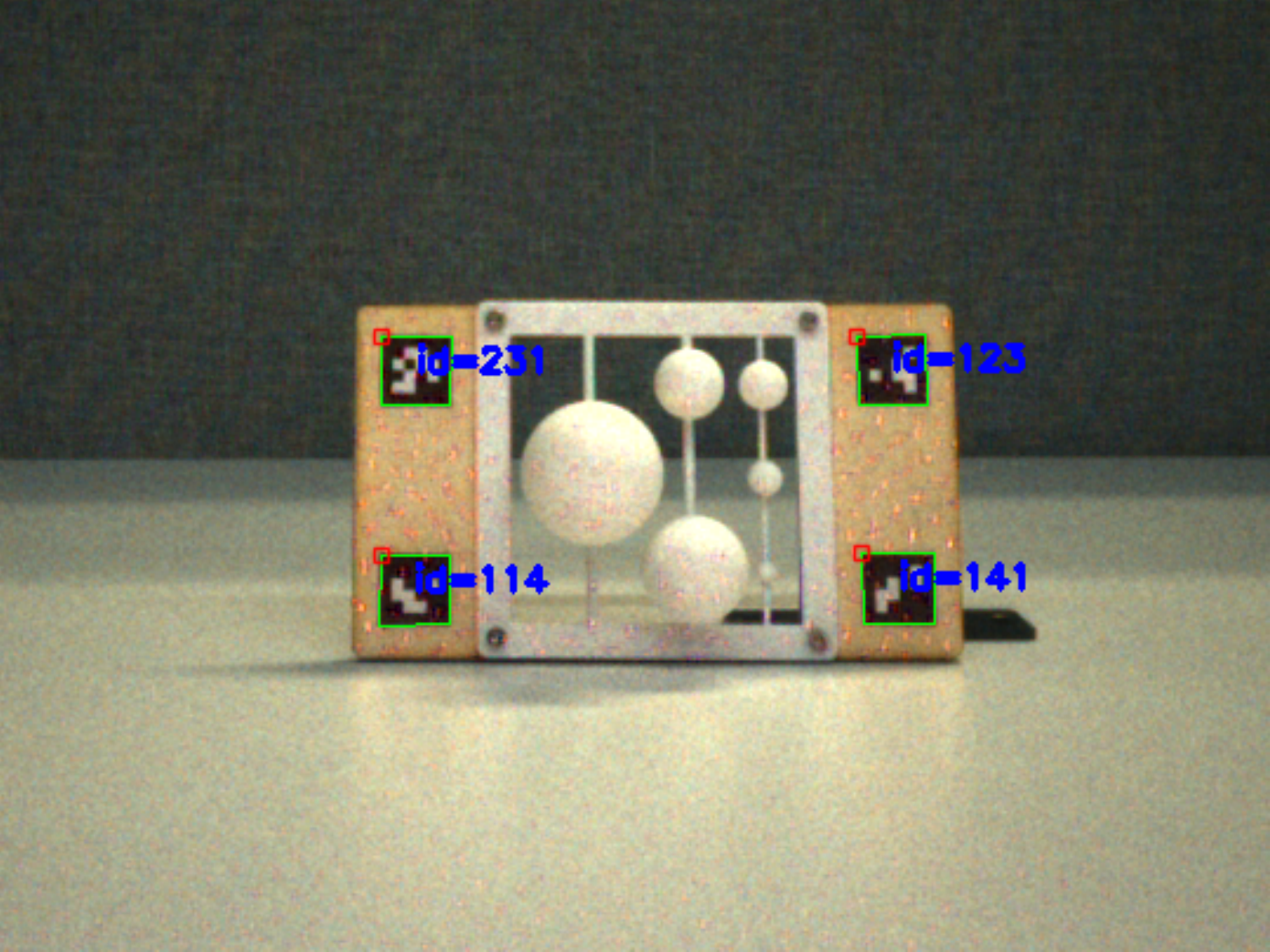}
}
\subfigure[]{	
\includegraphics[width=0.32\linewidth]{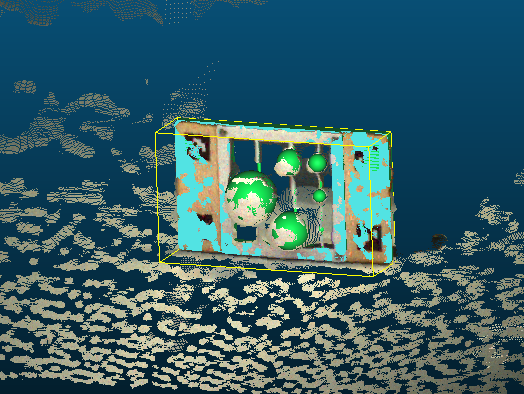}
}
}
\caption{A visualization of the alignment procedure: (a) An observed pointcloud before alignment; (b) The RGB image with the detected ArUco corners; (c) The pointcloud aligned with the reference mesh.}
\label{fig:pipeline}
\end{figure*}

\subsection{Fixture Design and Fabrication}

The test fixture is composed of 2 subassemblies, as shown in Fig. \ref{fig:assembly}. The lasercut backplate with attached fiducials is used for registration, and the 3D printed pattern sub-assembly holds the test pattern. The pattern subassembly consists of two parts: (1) the pattern plate, used to fixture the test pattern to the backplate, and (2) the test pattern itself.

We lasercut rather than 3D print the backplate to reduce manufacturing time, conserve 3D printer filament, and mark the locations of fiducials. To align the fiducials consistently, the square outlines for the fiducials (depicted as black squares in Fig. \ref{fig:assembly}) are etched into the backplate. The fiducials are printed on stickers and affixed to the backplate within the etched markings. In the experiments presented here, the backplate was cut out of $0.25$ inch birch plywood, but any flat material (of any thickness) can be used.

The same pattern plate is used for all test patterns, and can be imported into CAD or procedural modeling software to ensure that the test patterns fit into it. The pattern plate and test pattern subassembly is 3D printed in a single shot on a consumer-grade dual-filament FDM printer with support material.

FDM printing technology is perfectly suited for producing fixtures for measuring quality, since consumer-grade devices can print with sub-millimeter precision \cite{zhang_3d_2016}. Because the test fixture is printed directly from an existing 3D model, the reference mesh is guaranteed to be an accurate representation of the physical fixture.

To align and assemble the backplate and pattern plate, four $3$ mm holes are lasercut and printed respectively, intended to fit a screw-nut combination. To simplify fiducial marker detection, we use $20$ mm $\times$ $20$ mm markers, where the entire fixture is $173.6$ mm $\times$ $101.6$ mm.

With this paper we publish the CAD files necessary for lasercutting the backplate, 3D printing the test fixtures, and printing size-appropriate fiducials. The provided pattern plate 3D model can be used to generate (and then print) custom patterns.

\subsection{Alignment and Data Extraction}

For 3D registration, we use ArUco fiducial markers \cite{garrido-jurado_automatic_2014}, which can be detected under different orientation, perspective, skew, and lighting conditions. ArUcos can even be detected from grayscale images, allowing depth quality evaluation of not only RGBD but also grayscale stereo cameras. The downside of using ArUco markers is that all four corners need to be detected to have a valid high-confidence match.

Alignment to the test fixture is done by calculating a $4 \times 4$ rigid 3D transformation $H$ between the observed and actual 3D positions of the ArUco corners using the Umeyama method \cite{umeyama_least-squares_1991}. We need to compute both a rotation and translation component for alignment and registration. The deprojection equation to convert pixel location into 3D world coordinates is provided in Appendix \ref{sec:appendixalignment}. Using four ArUco tags on the backplate increases the robustness of the estimated alignment, since $16$ total points are used for the estimation. The estimated transformation is applied to the pointcloud to register it to the same coordinate system as the reference mesh.

We use a rigid instead of a full affine transformation to align the observed pointcloud with the reference mesh because we assume the cameras have no systemic source of skew. Robustly estimating a full affine transformation requires more fiducials that vary in depth, which adds complexity to the fixture manufacturing step. See Fig. \ref{fig:pipeline} for an example of an observed pointcloud, the detected ArUco tags, and the resulting aligned pointcloud.

\label{sec:alignment}

\subsection{Quality Analysis}
\label{sec:metric}

\begin{figure}[tpb]
\centering
\parbox{3.4in}{
\subfigure[]{
\includegraphics[width=0.46\linewidth]{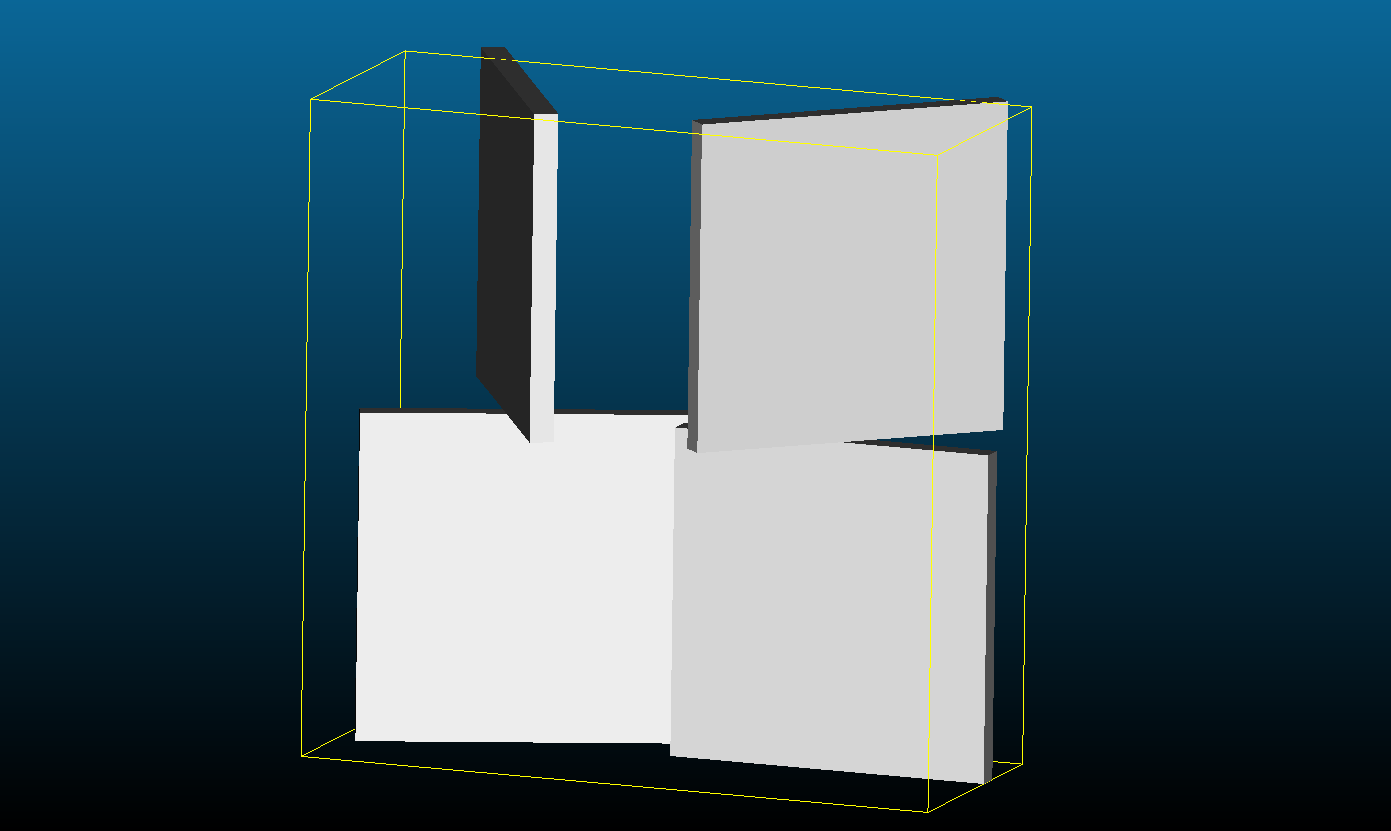}
}
\subfigure[]{
\includegraphics[width=0.46\linewidth]{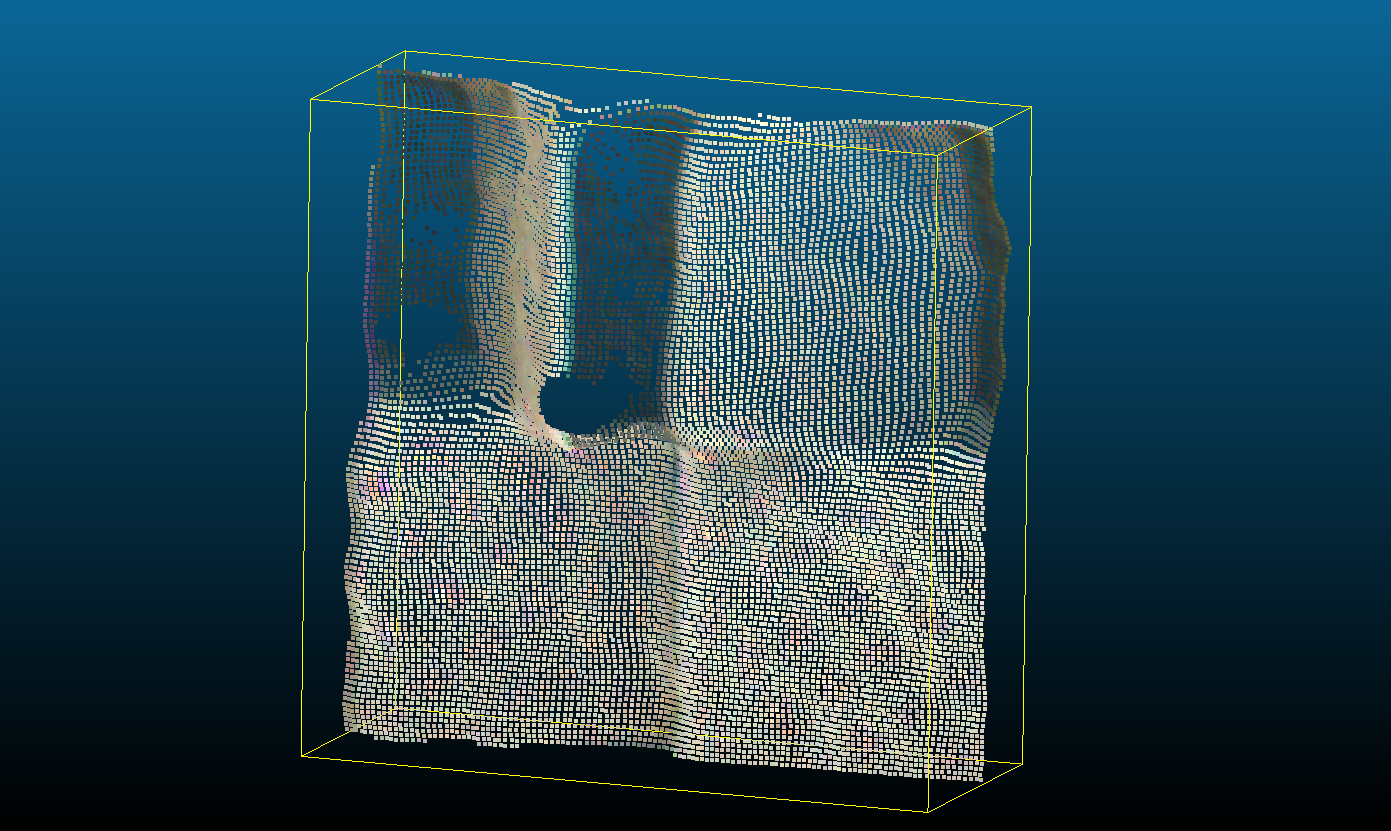}
}
}
\caption{Visual representation of the region of interest clipping. (a) A visualization of the region of interest; (b) an observed pointcloud after cropping to the region of interest.}
\label{fig:metricreg}
\end{figure}

Before calculating the RMSE and density of the observed pointcloud, we clip the aligned pointcloud to the area of interest. The area of interest is defined as the axis-aligned rectangular bounding box of the test pattern (without including the pattern plate) extended by a tolerance $t$ in all dimensions to capture points that are near the object at the boundary of the region of interest. Any points from the observed pointcloud that fall outside this volume are removed.

A visualization of the region of interest for the angled plates is shown in Fig. \ref{fig:metricreg} (a). The observed pointcloud at the backplate and pattern plate are not considered as part of the quality measurement because those areas are constant with every pattern type. An example cropped region can be seen in Fig. \ref{fig:metricreg}(b).

For each point $o_i$ in the cropped and aligned pointcloud, we compute the closest point $e_i$ on the reference mesh using libigl \cite{libigl}. RMSE is calculated by Eq. (1), where $n$ is the number of points in the cropped pointcloud. Ideally, each observed point will have no deviation from the closest point in the reference mesh.

$$
\text{RMSE} = \sqrt{\frac{1}{n}\sum^n_{i = 1}{(o_i - e_i)^2}} \eqno{(1)}
$$

Density is defined in pixels/$\text{mm}^2$ as the number of depth pixels that fall within a certain error tolerance $t$ per unit of visible fixture surface area, as shown in Eq. (2), where $A$ is the surface area of the mesh regions in the test pattern that are visible from the camera and $f_{\perp}$ is the face normal, shown in Eq. (3).

\[
\text{Density} = \frac{1}{A}\sum_{i = 1}^n \begin{cases} 1 \quad \text{if} \quad |o_i - e_i| < t \\ 0 \quad \text{otherwise} \end{cases} \eqno{(2)}
\]

\[
A = \sum_{f \in \text{faces}} \begin{cases} \text{Area}(f) \quad \text{if} \quad \arccos(f_{\perp} \cdot c) > \frac{\pi}{2} \\ 0 \quad \text{otherwise} \end{cases} \eqno{(3)}
\]

To compute the surface area of the reference mesh that is visible to the camera (Eq. (3)), we rotate the camera normal $(0, 0, -1)$ by the rotation component of the registration matrix $H$. From the mesh coordinate system, the camera normal is at $(0, 0, -1)$. The resulting vector is the camera normal $c$, representing the camera angle relative to the reference mesh. For each triangular face of the reference mesh, if the angle between the face normal and the camera normal is more than $\frac{\pi}{2}$ degrees, we consider it a face that has the potential to be visible from the camera.
 	
When analyzing results, each application will have different requirements on RMSE and density, so taking both metrics together on a set of different test patterns gives a clear understanding of depth quality for an RGBD camera.

\section{EXPERIMENTS}

\subsection{Camera Selection}

\begin{figure}[tpb]
\centering
\parbox{3.6in}{
\subfigure[]{
\includegraphics[width=.29\linewidth]{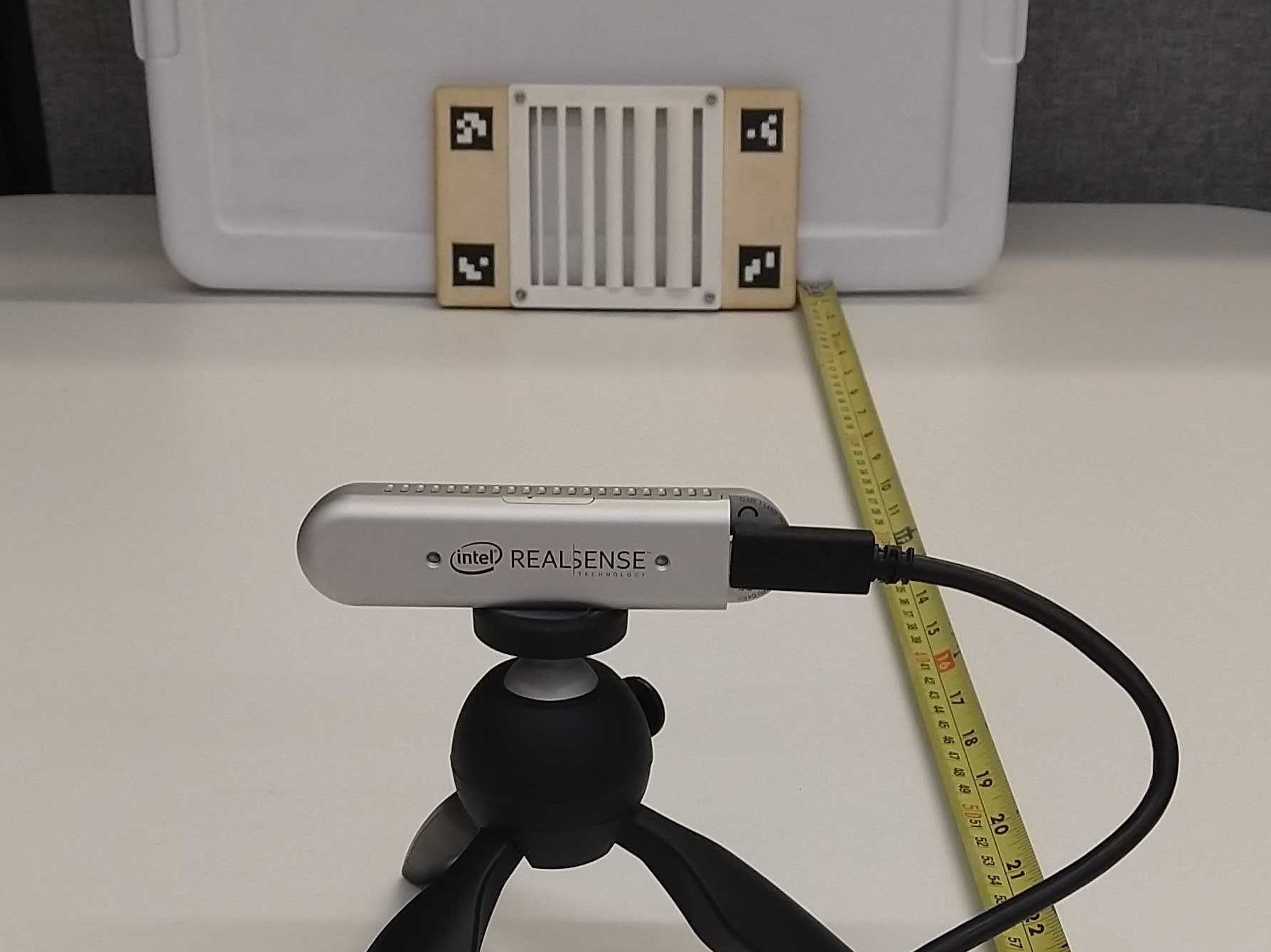}
}
\subfigure[]{
\includegraphics[width=.29\linewidth]{cylinder_reference_mesh}
}
\subfigure[]{
\includegraphics[width=.29\linewidth]{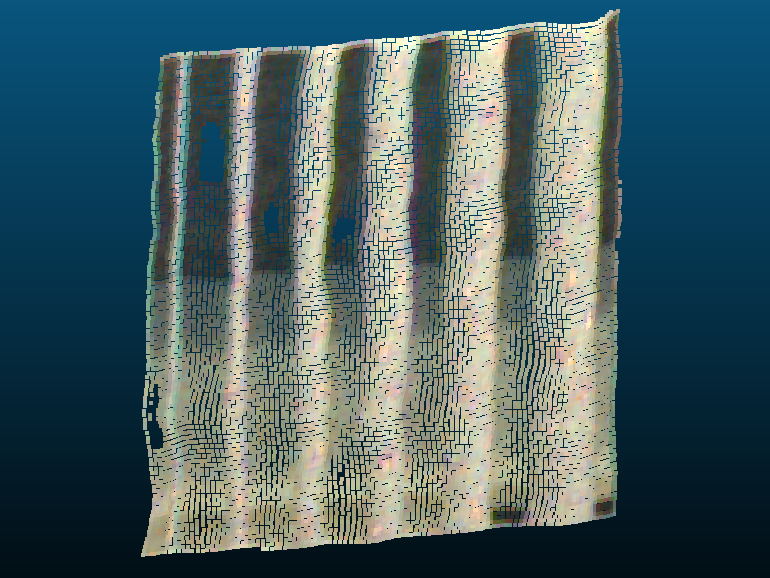}
}
}
\caption{The test setup for capturing Root AI camera selection data. (a) The camera and fixture. (b) The corresponding visualized reference mesh. (c) The observed pointcloud, cropped to relevant working area.}
\label{fig:test_setup}
\end{figure}

The three fixtures presented here can be used for camera selection. We compare 3 cameras (the Intel\textsuperscript{\textregistered} Realsense\textsuperscript{\texttrademark} D415, Intel\textsuperscript{\textregistered} Realsense\textsuperscript{\texttrademark} D435, and the ZED Mini) under a set of conditions for a hypothetical robotic manipulation use case with an expected operating distance of $24$ inches with an error tolerance $t$ of $0.002$ m using the test setup shown in Fig. \ref{fig:test_setup}. The Realsense\textsuperscript{\texttrademark} cameras are both active stereo cameras using an IR projector \cite{realsense_d400_datasheet}, and the ZED Mini camera is a passive stereo camera \cite{StereoLabs:ZedMini}. All three cameras use different optics, and have different different published quality metrics, but can still be compared with the metrics presented here.

The experiment was conducted by setting the test fixtures at $24$ inches from the front of the camera in the center of the camera field of view, to avoid any systematic error from field of view placement. No post-processing was done to the raw depth values, and the RMSE and density were calculated on 3 samples; the median values of each computed metric are repoted in Table \ref{tbl:camselection}. 

\begin{table}[h]
\caption{RMSE ($m$) and Density (pixels / $m^2$) for 3 candidate cameras}
\label{tbl:camselection}
\begin{center}
\begin{tabular}{|l||l|c|c|c|}
\hline
Fixture & Metric & D415 & D435 & ZED Mini \\
\hline
Cylinders & RMSE  & $\bm{0.00177}$ & $0.00200$ & $0.00319$ \\
& Density & $0.00144$ & $0.00137$ & $\bm{0.00197}$\\
\hline
Spheres & RMSE & $\bm{0.00269}$ & $0.00415$ & $0.00532$ \\
& Density      & $0.00150$ & $0.00098$ & $\bm{0.00182}$ \\
\hline
Angled plates & RMSE & $\bm{0.00223}$ & $0.00286$ & $0.00324$ \\
   & Density & $0.00145$ & $0.00140$ & $\bm{0.00223}$ \\
\hline
\end{tabular}
\end{center}
\end{table}

The D415 camera had the smallest RMSE for all 3 test fixtures at $< 2$ mm in error. The ZED Mini camera had the highest density for all 3 test fixtures, but nearly double the RMSE in all cases. For a manipulation application with an operating range of $24$ inches from the target, the D415 performs best.

\subsection{Horizontal vs. Vertical Cylinders}

\begin{figure}[tpb]
\centering
\parbox{3.6in}{
\subfigure[]{
\includegraphics[width=0.47\linewidth]{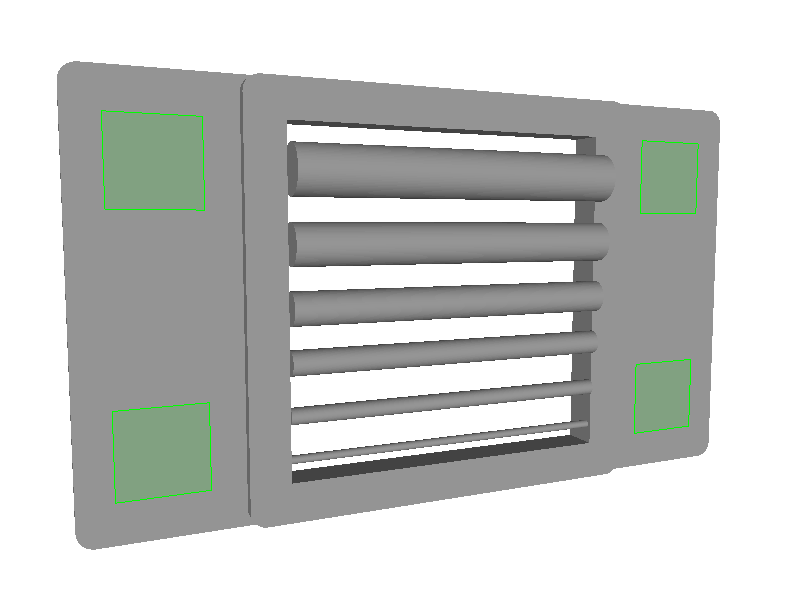}
}
\subfigure[]{
\includegraphics[width=0.47\linewidth]{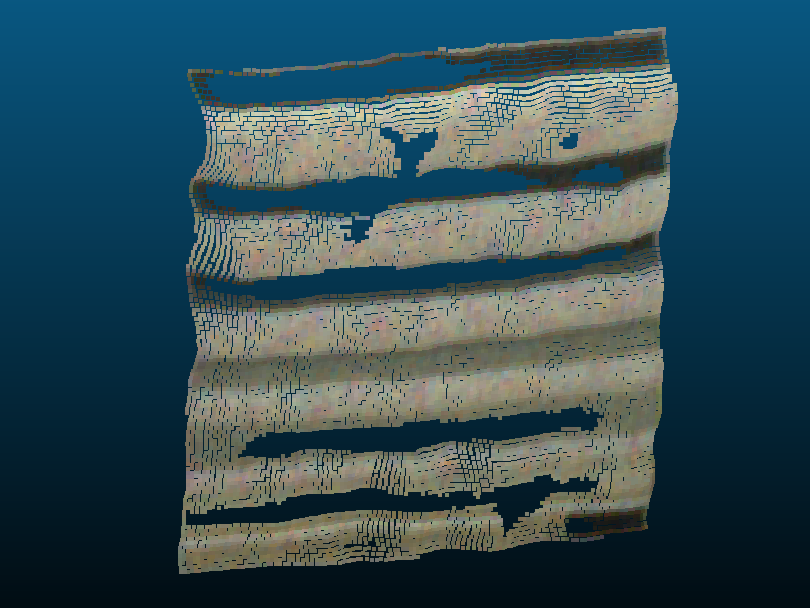}
}
}
\caption{(a) The reference mesh of horizontal cylinders. (b) An example cropped pointcloud for horizontal cylinders. Compared to the pointcloud in Fig. \ref{fig:pipeline} (c), this is less accurate.}
\label{fig:horiz_cylinders}
\end{figure}

Stereo matching is typically done by rectifying a pair of images and matching common pixels along horizontal epipolar lines \cite{scharstein_taxonomy_2001}. This means that for objects of uniform texture in the horizontal direction, like the horizontal cylinders in Fig. \ref{fig:horiz_cylinders} (a) stereo matching typically does poorly.

To test how much worse the performance is for horizontal and vertical lines, an Intel\textsuperscript{\textregistered} Realsense\textsuperscript{\texttrademark} D415 camera was used to capture 3 samples of both the horizontal and vertical cylinder reference meshes. The median RMSE and density is reported in Table \ref{tbl:cylinders}.

\begin{table}[h]
\caption{RMSE ($m$) and Density (pixels / $m^2$) for Horizontal and Vertical Cylinders}
\label{tbl:cylinders}
\begin{center}
\begin{tabular}{|l||c|c|}
\hline
Orientation & RMSE & Density \\
\hline
Horizontal & $0.00343$ & $0.00067$ \\
\hline
Vertical & $0.00177$ & $0.00144$  \\
\hline
\end{tabular}
\end{center}
\end{table}

The RMSE is 3 times worse and the density is 2 times lower for the horizontal cylinders, as expected. Using the modular fixture shown in Fig. \ref{fig:fixtures}, turning it into the horizontal cylinder fixture in Fig. \ref{fig:horiz_cylinders}(a) required simply unscrewing the fixture and screwing it back rotated by $90$ degrees, then generating a new reference mesh with the pattern plate and cylinder pattern rotated. 

\section{CONCLUSIONS}

The provided method can be extended by system integrators for running experiments with off-the-shelf RGBD cameras (or end-to-end RGBD camera systems) to determine optimal depth acquisition parameters, post-processing steps, environmental factors, and alternative fixtures. The provided reference meshes and CAD files\footnote{Meshes and code are available at: \url{http://github.com/Root-AI/depth-quality}} can be used for 3D printing custom fixtures that better approximate manipulation environments. To extend the provided code and models for custom fixtures, a new reference mesh needs to be generated for analysis.

The versatility and small size of these fixtures means that they can be taken into the field for depth quality calibration and testing as well.

Through a simple set of depth quality test fixtures that can easily be registered to a reference mesh, we can evaluate depth quality of RGBD cameras under any conditions.

\addtolength{\textheight}{-11.5cm}   



\section*{APPENDIX}

\subsection{Pointcloud to Mesh 3D Registration}
\label{sec:appendixalignment}

To calculate the 3D world coordinates of the ArUco corners, we take a depth image and deproject each pixel $(u, v, d)$ into a world coordinate $(x, y, z)$ according to the camera intrinsics and depth scale $s$ using the standard deprojection equation, as shown in Eq. (4), where $f_x$ and $f_y$ are the horizontal and vertical focal lengths and $(c_x, c_y)$ is the center of projection. The intrinsics are taken read directly from the camera and not calculated, since off-the-shelf depth cameras ship pre-calibrated.

\[
\begin{bmatrix}
x\\y\\z\\
\end{bmatrix} = 
\frac{d}{s}
\begin{bmatrix}
\dfrac{(u - c_x)}{f_x}\\
\dfrac{(v - c_y)}{f_y}\\
1 \\
\end{bmatrix}
 \eqno{(4)}
\]

The RGB color value at pixel location $(u, v)$ is the color used for the resulting world pixel. We rely on the implementation of Eq. (4) provided by each camera's publicly-available SDK.

\section*{ACKNOWLEDGMENT}

Thanks to the entire Root AI team for their support: Josh Lessing, CEO; Ryan Knopf, CTO; and Aayush Parekh, Mechanical Engineering Co-op.


\bibliographystyle{./IEEEtran}
\bibliography{IEEEabrv,references}

\end{document}